%% file: main.tex
\documentclass[10pt,twocolumn,letterpaper]{article}

\usepackage{cvpr}  
\usepackage{graphicx} 
\usepackage{xcolor}
\usepackage{colortbl}
\usepackage{enumitem}

\usepackage[pagebackref,breaklinks,colorlinks]{hyperref}

\usepackage[capitalize]{cleveref}
\crefname{section}{Sec.}{Secs.}
\Crefname{section}{Section}{Sections}
\Crefname{table}{Table}{Tables}
\crefname{table}{Tab.}{Tabs.}

\newcommand{\tablestyle}[2]{\setlength{\tabcolsep}{#1}\renewcommand{\arraystretch}{#2}\centering\footnotesize}
\newlength\savewidth\newcommand\shline{\noalign{\global\savewidth\arrayrulewidth
		\global\arrayrulewidth .8pt}\hline\noalign{\global\arrayrulewidth\savewidth}}

\definecolor{maroon}{cmyk}{0,0.1,0.01,0.01}
\definecolor{blue}{cmyk}{0.95,0.0,0.2,0.2}
\definecolor{yellow}{cmyk}{0.01,0.0,0.2,0.01}
\definecolor{lightblue}{cmyk}{0.1,0.0,0.02,0.02}
\definecolor{darkgreen}{RGB}{51,181,41}
\definecolor{darkorange}{RGB}{252,135,62}
\definecolor{lightgreen}{HTML}{f1f2e4}
\definecolor{lightgray}{gray}{0.95}

\begin{document}

\title{VLMEvalKit: An Open-Source Toolkit\\for Evaluating Large Multi-Modality Models}

\author{Haodong Duan \hspace{3mm}  Xinyu Fang \hspace{3mm} Junming Yang \hspace{3mm} Xiangyu Zhao \hspace{3mm} Zerun Ma \hspace{3mm}  Yuxuan Qiao\\ 
Mo Li \hspace{3mm} Tianhao Liang \hspace{3mm} Lin Zhu\hspace{3mm} Amit Agarwal \hspace{3mm}  Xiaozhe Li \hspace{3mm} Shengyuan Ding \hspace{3mm}\\
Jiazi Bu* \hspace{3mm} Ziyu Liu* \hspace{3mm} Zhangyang Qi* \hspace{3mm} YiFei Li* \hspace{3mm} Yuhang Zang* \hspace{3mm} \\
Zhe Chen \hspace{3mm} Lin Chen \hspace{3mm} Yuan Liu \hspace{3mm} Yubo Ma \hspace{3mm} Hailong Sun \hspace{3mm} Yifan Zhang \hspace{3mm} Shiyin Lu \hspace{3mm}\\
Tack Hwa Wong \hspace{3mm} Weiyun Wang \hspace{3mm} Peiheng Zhou \hspace{3mm} Chaoyou Fu \hspace{3mm}Junbo Cui \hspace{3mm} Jixuan Chen   \\
Enxin Song \hspace{3mm} Song Mao \hspace{3mm} Junming Lin \hspace{3mm} Xilin Wei \hspace{3mm} Jinsong Li \hspace{3mm} Zeyi Sun\hspace{3mm} Zhaowei Wang  \\
Zicheng Zhang \hspace{2mm} Xiaoyi Dong \hspace{2mm} Junjun He \hspace{2mm}  Pan Zhang \hspace{2mm} Jiaqi Wang \hspace{2mm} \\
Dahua Lin \hspace{2mm} Kai Chen 
\vspace{3mm} \\
$^1$VLMEvalKit Team \hspace{5mm} $^2$Community Contributors
}

\maketitle

\input{sections/abstract}
\input{sections/intro}
\input{sections/design}

\input{sections/evaluation}

\input{sections/discussion}

{
\small
\bibliographystyle{ieee_fullname}
\bibliography{egbib}
}

\clearpage
\input{sections/appendix}

\end{document}

%% file: sections/abstract.tex
\begin{abstract}
We present VLMEvalKit: an open-source toolkit for evaluating large multi-modality models based on PyTorch. 
The toolkit aims to provide a \textbf{user-friendly} and \textbf{comprehensive} framework
for researchers and developers to evaluate existing multi-modality models and publish \textbf{reproducible} evaluation results.
In VLMEvalKit, we implement over \textbf{450+ large multi-modality model configurations}, including both proprietary APIs and open-source models,
and support \textbf{330+ benchmark} across diverse multi-modal benchmarks.
By implementing a single interface, new models can be easily added to the toolkit, 
while the toolkit automatically handles the remaining workloads, including data preparation, distributed inference, prediction post-processing, and metric calculation. 
VLMEvalKit has also evolved to a broader evaluation suite spanning video/audio, document understanding, GUI grounding, spatial reasoning, safety, scientific reasoning, and multi-turn dialogue.
Based on the evaluation results obtained with the toolkit, 
we host \href{https://huggingface.co/spaces/opencompass/open_vlm_leaderboard}{OpenVLM Leaderboard},
a comprehensive leaderboard to track the progress of multi-modality learning research.
The toolkit is released on 
\href{https://github.com/open-compass/VLMEvalKit}{GitHub}
and is actively maintained\footnote{
VLMEvalKit contributors can join the author list of the report based on their contribution to the repository. Specifically, it requires 3 major contributions (implement a new benchmark, MLLM, or contribute a major feature). 
We will update the report quarterly and an additional section that details each developer's contribution will be appended in the next update. \textbf{* Donates special thanks for the report updating.}
}.
\end{abstract}

%% file: sections/intro.tex
\section{Introduction}

With the rapid development of Large Language Models (LLMs) 
\cite{2022chatgpt,touvron2023llama,2023internlm,Claude3},
Large Multi-Modality Models (LMMs)~\cite{OpenAI2023GPT4TR,team2023gemini} have also experienced significant advancements. 
LMMs typically take two or more modalities as input. 
Most of the research has focused on LMMs for image and text~\cite{liu2023visual,chen2023sharegpt4v}, 
but research has also been extended to other modalities, such as audiotext~\cite{chu2023qwen}, video~\cite{song2024moviechat,li2024mvbench,chen2024sharegpt4video}, 
or point clouds~\cite{xu2023pointllm,xue2024ulip}.
Furthermore, there exist LMMs that can simultaneously take more than two modalities as inputs, including proprietary APIs and open-source models~\cite{team2023gemini,OpenAI2023GPT4TR,han2024onellm}.
Compared to previous multi-modality models, 
LMMs, empowered by large language models, exhibit enhanced generalization capability and engage with humans in a variety of conversational styles.
These models have not only demonstrated remarkable capabilities in multi-modal perception and reasoning tasks but have also spurred a range of innovative applications.

Quantitative evaluation is crucial in the development of LMMs. 
As general-purpose models, 
LMMs must undergo rigorous evaluation in a diverse range of tasks and domains. 
Comprehensive evaluations not only help users discern the strengths and weaknesses of an LMM, 
but also offer valuable feedback to developers for ongoing refinement. 
Unlike `pre-GPT' models, 
evaluating LMMs on diversified quantitative benchmarks has become a common practice, 
both in academic works~\cite{liu2024llavanext,ye2023mplugowl2} and commercial APIs~\cite{OpenAI2023GPT4TR,team2023gemini}.

Despite the importance of evaluating LMMs,
conducting assessments across dozens of benchmarks can be a daunting task, 
particularly for small research teams. 
One must prepare data based on numerous repositories and manage potential environmental conflicts.
Moreover, the authors of benchmarks may not provide evaluation results for all LMMs in which users are interested, 
thus requiring significant effort to compile uncompleted results.
To alleviate this challenge, we developed VLMEvalKit, 
an open-source toolkit designed to facilitate the evaluation of LMMs.

\begin{figure}[t]
\centering
\includegraphics[width=\linewidth]{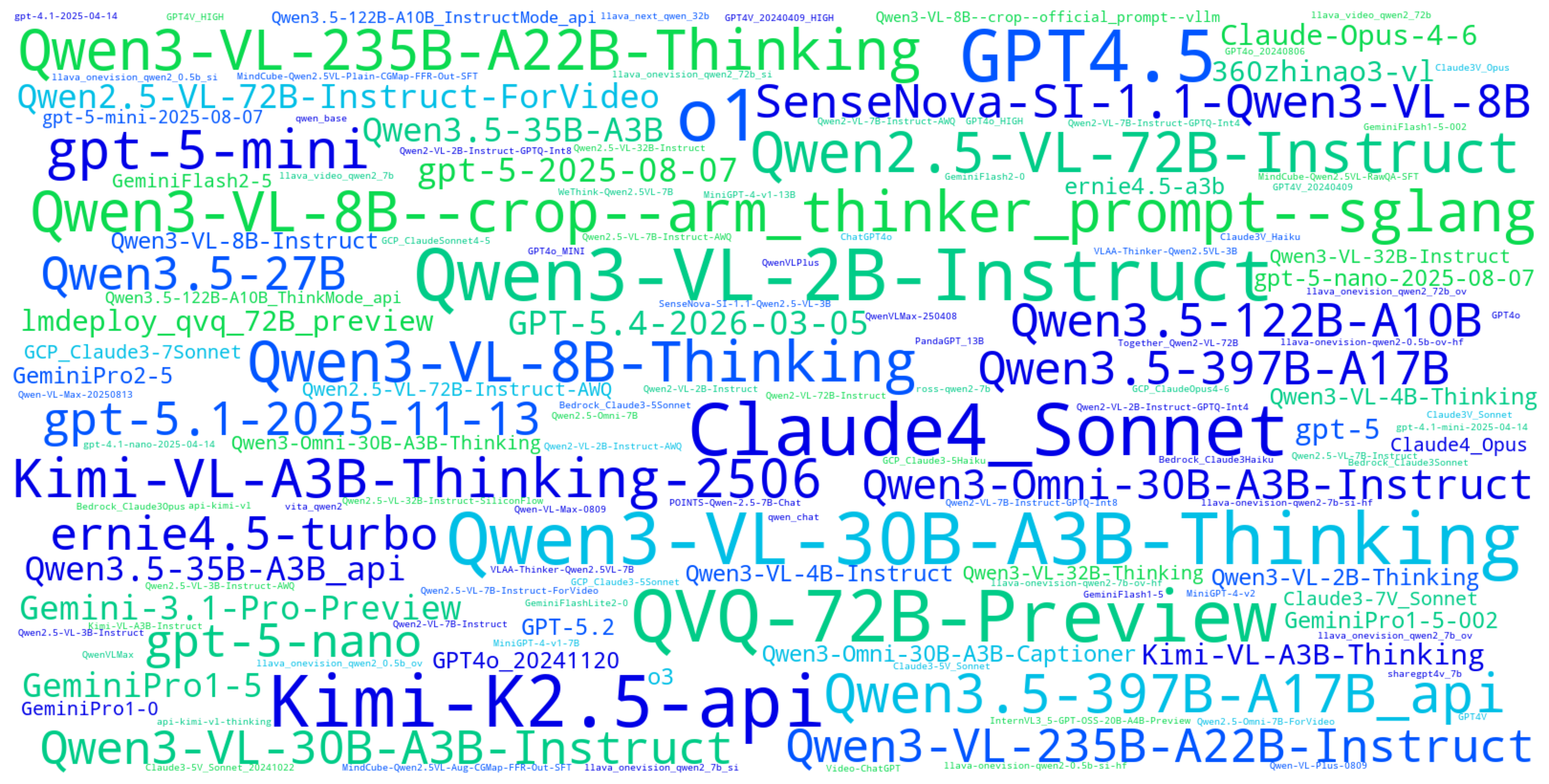}
\vspace{-5mm}
\caption{VLMEvalKit supports most mainstream commercial APIs including GPT-5, Gemini-3.1, Claude-opus-4.6, \emph{etc.}, as well as over 300 open-source LMMs including LLaVA~\cite{liu2023visual}, InternVL~\cite{chen2024far}, Qwen~\cite{Qwen2.5-VL}, \emph{etc.}}
\vspace{-4mm}
\label{fig:supported_models}
\end{figure}

VLMEvalKit aims to provide a comprehensive, user-friendly evaluation framework for 
researchers and developers to assess existing Large Multi-Modality Models. 
Currently, the codebase supports more than 450 different large multi-modality models,  
spanning proprietary APIs and open-source models (see \cref{fig:supported_models}), 
and over 335 multi-modal benchmarks covering a wide range of tasks and scenarios,
with representative benchmarks shown in \cref{tab:supported_benchs}.
Its evaluation scope has expanded from image-text VLM benchmarks to video understanding, document/OCR/chart analysis, GUI grounding, spatial reasoning, safety evaluation, medical and scientific reasoning, generation, and multi-turn dialogue scenarios.
The codebase's straightforward design simplifies the integration of new benchmarks or LMMs.
Typically, a developer only needs to prepare a single data file or implement a single interface 
to support new benchmarks or LMMs.
Beyond its extensive collection and simplified design, 
VLMEvalKit significantly eases the work of comprehensive evaluation. 
Users can launch evaluations across multiple supported LMMs and benchmarks 
with a single command, generating well-structured evaluation results. 
The entire process eliminates the need for manual data preparation or post-processing, 
ensuring a seamless and efficient evaluation experience.

VLMEvalKit Lite version\footnote{
https://github.com/OpenEvaluation/VLMEvalKit
} further complements this workflow with a lightweight tooling layer for local inspection and result analysis. 
It provides Flask-based interfaces for browsing datasets, inspecting the \texttt{build\_prompt()} output of individual samples, reviewing prediction and judge files together with the original visual inputs, and loading summarized results into a local leaderboard. 
Together with reorganized tutorials and troubleshooting-oriented documentation, these tools make VLMEvalKit easier to operate beyond batch execution, especially when users need to debug new benchmarks, verify model prompts, or communicate evaluation results.

To align with real use cases, 
VLMEvalKit employs generation-based evaluation across all LMMs and benchmarks. Given their general-purpose nature, 
LMMs may struggle with adhering to specific formats for solving multiple-choice questions.
For a fair comparison, VLMEvalKit utilizes large language models as choice extractors when exact matching fails, 
thereby mitigating the impact of response styles and enhancing the reliability of evaluation, 
Leveraging the reliable and reproducible evaluation results provided by VLMEvalKit, 
we maintain a comprehensive leaderboard to monitor the advancement of LMM development. 
The leaderboard disseminates valuable insights to the community and has garnered widespread recognition from both academia and industry.

VLMEvalKit is publicly available at \url{https://github.com/open-compass/VLMEvalKit} under the Apache 2.0 License.
The repository includes the complete source codes along with detailed instructions for installation, evaluation, and further development. 
In subsequent sections,
we will delve into the design and features of VLMEvalKit, 
present and analyze the evaluation results obtained, 
and discuss potential future developments.

\begin{table*}[t]
\centering
\resizebox{\linewidth}{!}{
\tablestyle{4pt}{1.2}
\begin{tabular}{l|p{14cm}}
\shline
\textbf{Category} & \textbf{Representative Benchmarks} \\
\shline
\textbf{Video \& Omni} & Video-MME-v2, VideoMMMU, Video-TT, MMSI-Video-Bench, MVU\_Eval, VSI-Bench, DREAM-1K, VTCBench, VideoZeroBench \\
\hline
\textbf{Document, OCR \& Chart} & SciDocBench, OCRBench\_v2\_MINI, olmOCRBench, OceanOCRBench, MAT-Bench, ChartMuseum, ChartQAPro, ChartBench, ChartX, ChartCap, PlotQA, Design2Code \\
\hline
\textbf{Spatial, GUI \& Grounding} & OSWorld-G, TopViewRS, RefCOCO, GroundingME, DA-2K, ERQA, RefSpatialBench, VenusBench-GD, VBGD, WorldVQA, UniSVG \\
\hline
\textbf{Safety \& Bias} & VLMBias, MMSafetyBench, XSTest, MSSBench, Flames, SIUO, M3oralBench, VLMs Are Biased \\
\hline
\textbf{Medical \& Scientific} & MedQ-Bench, Asclepius, MMESCI, MMOral-OPG, NPMM-bench, SGI-Bench 1.0, MMReason, HiPhO, CoreCognition, SSI-Bench \\
\hline
\textbf{Comprehensive \& Multi-Task} & MMBench, MMMU, SEEDBench2-Plus, MME, LLaVABench \\
\shline
\end{tabular}}
\vspace{-2mm}
\caption{VLMEvalKit supports the evaluation of over 335 multi-modal benchmarks, spanning from conventional image-text comprehension to specialized domains including Video, Document/Chart analysis, Spatial \& GUI grounding, Safety, and Scientific reasoning. }
\label{tab:supported_benchs}
\end{table*}

%% file: sections/design.tex
\section{VLMEvalKit: Design \& Features}

\begin{figure}[t]
\centering
\includegraphics[width=.9\linewidth]{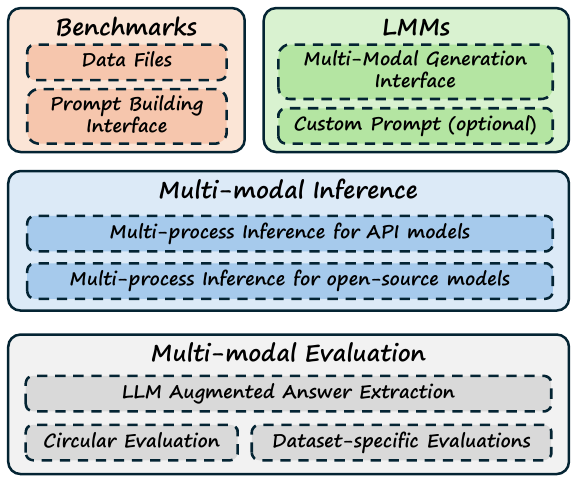}
\vspace{-3mm}
\caption{Major Components in VLMEvalKit. }
\vspace{-4mm}
\label{fig:design}
\end{figure}

\cref{fig:design} displays the major components in VLMEvalKit. 
In this section, we will describe each single component in detail.

\noindent
\textbf{Benchmarks. }
% VLMEvalKit encompasses dozens of multi-modal benchmarks that span a broad spectrum of tasks and scenarios, as detailed in \cref{tab:supported_benchs}. 
VLMEvalKit encompasses over 335 multi-modal benchmarks that span a broad spectrum of tasks and scenarios, as detailed in \cref{tab:supported_benchs}.
The benchmarks are first processed into \texttt{.tsv} files, 
where evaluation samples are stored as individual lines.
Each evaluation sample typically includes the \texttt{index}, \texttt{question}, \texttt{answer}, \texttt{image} (one or more images encoded in \texttt{base64}), 
and \texttt{choices} for multi-choice questions. 
Each dataset class supports a \texttt{.build\_prompt()} interface, 
which the evaluation script utilizes to construct evaluation samples into multi-modal messages. 
A multi-modal message comprises an interleaved sequence of contents across various modalities. 
For illustration, consider the following example:

\noindent
\texttt{\small [ \\ 
\hspace*{5mm} dict(type='image', value='path or url of the image'), \\
\hspace*{5mm} dict(type='image', value='path or url of the image'), \\
\hspace*{5mm} dict(type='text', value='Please list all the objects that appear in the above images.') \\
]}

Currently, a multi-modal message primarily incorporates \texttt{image}, \texttt{video} and \texttt{text} modalities.
The format, however, is extensible and can accommodate additional modalities such as audio or point clouds.

Beyond conventional image-text evaluation, the toolkit has significantly expanded its scope to accommodate the rapid evolution of multi-modality learning. We have introduced comprehensive support for specialized domains, including Document/OCR/Chart analysis (e.g., SciDocBench, ChartX), Safety \& Bias evaluation, and Medical/Scientific reasoning (e.g., MedQ-Bench, SGI-Bench).

Furthermore, keeping pace with the latest trends in Agentic AI and Omni-modal understanding, VLMEvalKit now provides robust evaluation pipelines for Spatial/GUI grounding and Video understanding. Benchmarks such as OSWorld-G, DA-2K, Video-MME-v2, and VideoMMMU are fully integrated, enabling developers to rigorously evaluate LMMs on complex, dynamic agentic environments and long-context video tasks. %Simple Tool can also be invoked by models via 

\noindent
\textbf{LMMs. }
% VLMEvalKit supports over 200 LMMs, including both commercial APIs and open-source models. 
VLMEvalKit supports over \textcolor{red}{450} LMMs, including both commercial APIs (\eg, GPT, Gemini, and Claude) and open-source models (\eg, InternVL series, QwenVL series, and Kimi). 
A unified \texttt{.generate()} interface has been implemented for all LMMs, 
which accepts a multi-modal message as input and returns the response string.
For LMMs that are limited to processing a single image-text pair, 
the concatenated text messages and the first image are adopted as the default input.
To provide flexibility for LMM developers,
the \texttt{.generate()} interface also includes \texttt{dataset\_name} as an additional argument. 
An LMM can optionally implement its own \texttt{.build\_prompt()} interface to construct custom multi-modal messages, 
and determine whether to utilize custom multi-modal messages based on the \texttt{dataset\_name} flag.
The unified interface eases the process of comprehensive evaluation. 
Each newly supported LMM / benchmark can be directly evaluated with all existing benchmarks / LMMs.

\noindent
\textbf{Multi-modal Inference. }
To expedite the multi-modal inference,
we support parallelized inference for both commercial APIs,
leveraging Python's \texttt{multiprocessing},
and open-source models, which are distributed in parallel across multiple GPUs.
For some popular open-source LMMs (like Qwen Series, InternVL Series), we have supported vLLM / LMDeploy to accelerate the inference with full utilization of resources.
Furthermore, VLMEvalKit adopts a unified evaluation entry point: the previously separate \texttt{run\_api.py} has been merged into \texttt{run.py} through an \texttt{--api-mode} flag, so that API-based and locally hosted models share the same command-line interface, configuration path, and judge-model arguments. The same launch also supports sequential inference across all datasets and parallel evaluation, allowing a model to be assessed over many benchmarks with a single command while the comparatively lightweight scoring stage runs concurrently.

To make long evaluation campaigns both robust and cost-efficient, the reuse mechanism has been overhauled. Each run now records a structured status file that tracks per-dataset progress across the \texttt{pending} / \texttt{infer} / \texttt{eval} / \texttt{done} stages, so that an interrupted run can be resumed with format-aware reuse of existing prediction files and minimal repeated computation or API calls. A three-level \texttt{--reuse-aux} control (\texttt{all} / \texttt{infer} / \texttt{none}) further lets users decide how aggressively intermediate artifacts are reused, while standardized checkpoint keys ensure consistent serialization across image, multi-turn, and video inference paths.

\input{sections/vqa_main_table}

\input{sections/reasoning_main_table}

\noindent
\textbf{Tooling / Lite version. }
Beyond the core evaluation pipeline, VLMEvalKit Lite provides some lightweight tools for inspecting data, debugging evaluation artifacts, and publishing local results. 
In the Lite branch, the existing dataset browser is refactored into a Flask-based web tool, while the evaluation result browser, leaderboard, and tool hub are added as companion Flask applications. 
These web tools expose blueprint registration functions, so they can be launched as standalone applications or mounted together under separate URL prefixes in a local hub. 
This design keeps the one-command evaluation workflow unchanged, while giving users interactive entry points to examine the same datasets, prompts, predictions, and score files that are consumed by \texttt{run.py} and dataset-level \texttt{evaluate()} / \texttt{report()} interfaces.

The dataset browser loads benchmarks through \texttt{build\_dataset()} and presents samples in a paginated table. 
It renders inline image thumbnails with an image carousel for multi-image samples, supports page jumping, and exposes full row details after selection. 
For efficient dataset auditing, it automatically detects category-like columns and supports category filtering as well as substring search over selected fields. 
Most importantly for model integration, the browser calls \texttt{build\_prompt(line)} for the selected sample and visualizes the resulting multi-modal message, making it easier to verify image materialization, interleaved image-text ordering, and dataset-specific prompt templates before large-scale inference.

The Lite Version also includes an evaluation result browser and a Flask-based leaderboard. 
The result browser scans an output root to discover prediction, judge, and rating files for a given model-dataset pair, then maps each result row back to the original dataset by \texttt{index} so that predictions can be inspected together with the source images. 
It offers the same pagination, category filters, substring search, and row-level detail view, while hiding large raw image fields and emphasizing result columns such as prediction, answer, judge output, and hit status. 
The leaderboard reads summarized evaluation JSON from a local path, a web URL, or an uploaded file, enabling users to review local or shared results without depending on a hosted service.

The Lite branch further reorganizes the documentation around the practical evaluation lifecycle. 
The documentation index points users from quick start to a complete tutorial, then to workflow, model interfaces, dataset conventions, tools, environment setup, and troubleshooting; the documentation set also includes VLLM deployment notes. 
The tutorial uses a concrete Seed2.0-Lite on MMMU\_DEV\_VAL example to connect model and dataset discovery, \texttt{run.py} execution, output directory conventions, judge artifacts, and result reading through \texttt{report()}. 
Together with recent video benchmark refactoring such as VideoZeroBench integration, these additions turn VLMEvalKit from a unified backend for reproducible evaluation into a more inspectable and operator-friendly evaluation workbench.

\noindent
\textbf{Multi-modal Evaluation. }
Predictions from the LMMs will be evaluated based on the specific question format to derive the final metrics. 
Benchmarks supported in VLMEvalKit can be categorized into three primary types: 
1. Multi-choice questions (MCQ), 
where the model must select from given options and respond with the corresponding label (\emph{e.g.}, A, B);
2. Yes-or-No questions (Y/N),
requiring a straightforward `Yes' or `No' answer;
3. Open-ended questions, which necessitate a free-form response. 
Notably, a significant number of LMMs struggle to adhere to the instructions precisely, 
often producing responses that are not well-formatted for MCQ and Y/N benchmarks.
To improve the precision of evaluations and counteract the influence of varied response styles,
VLMEvalKit offers the option to integrate LLM-augmented answer extraction
specifically for MCQ and Y/N benchmarks. 
We first adopt exact matching to match the response with the option labels or contents.
Should this step fail, the toolkit then prompts an LLM (such as ChatGPT) to match the response with the option that most closely aligns with semantic meaning.
This strategy helps us better understand the real performance of LMMs, particularly for commercial APIs~\cite{liu2023mmbench}.

Another notable challenge in assessing MCQ benchmarks is the inherent \textbf{variance}. 
When employing random guessing, an LMM may correctly answer $1 / N$ questions for $N$-option multi-choice benchmarks. 
Besides, we find that the outcome of an LMM can be significantly influenced by the order of options. 
These factors make the evaluation results highly variable and the performance gap between LMMs less discernible. 
To address this, VLMEvalKit offers an option to evaluate all MCQ benchmarks in \textbf{Circular} mode:
$N$ options of a MCQ will be shifted in circular $N$ times to formulate $N$ new questions.
The results count only if an LMM accurately answers all $N$ circular-shifted MCQs.
The CircularEval strategy can more effectively assess the real comprehension of an LMM on MCQ benchmarks,
allowing users to identify more pronounced performance disparities between models.

For open-ended benchmarks, VLMEvalKit follows the original practice for conducting the evaluation. 
Subjective benchmarks like MMVet~\cite{yu2023mm} or LLaVABench~\cite{liu2023visual} adopts GPT-4 for marking,
based on the semantic similarity between LMM responses and the reference answer.
For VQA benchmarks~\cite{mishra2019ocr,textvqa},
we follow the standard practices to calculate accuracies based on heuristic matching.

% As LMMs increasingly produce long responses, VLMEvalKit has extended its open-ended evaluation toward reasoning-aware scoring. For benchmarks such as SciDocBench, the assessment of the final answer is decoupled from the assessment of the intermediate reasoning process, so that answer correctness and reasoning quality are reported as separate signals rather than conflated into a single score. The GPT-assisted scoring pipeline has likewise been made more robust: judge-based benchmarks (\eg, XLRSBench) share a common \texttt{build\_judge()} interface, run answer extraction and scoring in parallel with a resumable intermediate file, and gracefully fall back to deterministic regex / exact matching when no judge model is configured.

VLMEvalKit also broadens evaluation beyond single-turn settings and hardens the evaluators themselves. Multi-turn chat is now supported for most commercial API models, with the multi-turn datasets refactored onto a shared conversation interface, so that dialogue-style benchmarks reuse the same inference and scoring path as single-turn ones. For evaluators that execute model-generated artifacts or invoke external judges---such as the chart-understanding benchmarks ChartX and ChartMimic---the toolkit adds timeout guards (\eg, a 600s per-item limit), broad exception handling that returns a zero score instead of aborting the whole run, and a series of answer-extraction bug fixes, making large-scale evaluation substantially more stable.

%% file: sections/vqa_main_table.tex
\begin{table*}[t]
\centering
\resizebox{\linewidth}{!}{
\tablestyle{6pt}{1.5}
\begin{tabular}{ccccccccccc}
\shline
\rowcolor{lightgray}
Method                      & Param. & Avg. Score & MMBenchV1.1 & MMStar & MMMU & Math. & Hallu. & AI2D & OCR. & MMVet \\
\shline
\rowcolor{lightblue}
SenseNova-V6-5-Pro~\cite{SenseNova}  & N/A    & \textbf{82.2} & 87.3 & 76.1 & 77.0 & \textbf{82.8} & \underline{66.7} & \textbf{90.2} & 885 & \textbf{89.4} \\
\rowcolor{lightblue}
CongRong-v2.0~\cite{CongRong-v2}     & N/A    & \underline{80.7} & \underline{88.1} & 75.3 & 75.6 & 76.8 & 63.2 & \underline{90.0} & \underline{927} & 83.9 \\
\rowcolor{lightblue}
SenseNova-V6-Pro~\cite{SenseNova}    & N/A    & 80.4 & 88.0 & 73.7 & 70.4 & 76.9 & \textbf{67.1} & 89.2 & 895 & \underline{88.2} \\
\rowcolor{lightblue}
Gemini-2.5-Pro~\cite{team2023gemini} & N/A    & 80.1 & \textbf{88.3} & 73.6 & 74.7 & 80.9 & 64.1 & 89.5 & 862 & 83.3 \\
\rowcolor{lightblue}
GPT-5-20250807~\cite{2022chatgpt}    & N/A    & 79.9 & 86.6 & 75.7 & \textbf{81.8} & 81.9 & 65.2 & 89.5 & 807 & 77.6 \\
\rowcolor{lightblue}
JT-VL-Chat-V3.0                      & N/A    & 79.9 & 87.5 & \textbf{82.1} & 68.7 & 72.8 & 64.4 & 88.3 & \textbf{950} & 80.3 \\
InternVL3-78B~\cite{zhu2025internvl3}& 78B    & 79.1 & 87.7 & 73.4 & 72.2 & 79.0 & 59.1 & 89.8 & 908 & 80.7 \\
BlueLM-2.6-3B                        & 3B     & 78.4 & 86.4 & \underline{80.1} & 62.4 & \underline{82.3} & 63.1 & 86.1 & 881 & 78.5 \\
\rowcolor{lightblue}
GPT-5-mini-20250807~\cite{2022chatgpt}& N/A   & 78.0 & 86.2 & 73.2 & \underline{78.7} & 79.2 & 62.5 & 86.7 & 828 & 74.6 \\
InternVL3-38B~\cite{zhu2025internvl3}& 38B    & 77.8 & 86.8 & 72.6 & 69.7 & 76.3 & 58.4 & 88.7 & 886 & 81.1 \\
\shline
\end{tabular}}  
\vspace{0.5mm}
\caption{The evaluation results of LMMs on general VQA benchmarks. 
The table displays the top-10 LMMs, including both commercial APIs and open-source LMMs(\textbf{till 2025.09.17}), in the descending order of average score. 
When calculating the average score, 
scores of each benchmark are normalized to the range of 0 to 100.
When reporting the parameter size, `1B' means $10^9$ parameters. 
Commercial APIs are denoted with \colorbox{lightblue}{blue background}. 
\textbf{Bold} and \underline{underline} indicates the best and second-best performance in each group.
}
\label{tab:vqa_main_results}
\vspace{-4mm}
\end{table*}

%% file: sections/reasoning_main_table.tex
\begin{table*}[t]
\centering
\resizebox{\linewidth}{!}{
\tablestyle{6pt}{1.5}
\begin{tabular}{ccccccccccc}
\shline
\rowcolor{lightgray}
Method                      & Param. & Avg. Score & MathVista & MathVision & MathVerse & DynaMath & WeMath & LogicVista  \\
\shline
\rowcolor{lightblue}
Seed1.5-VL~\cite{guo2025seed15vl}  & N/A & 73.3 & \textbf{86.8} & 67.3 & \underline{79.3} & 56.1 & \underline{77.5} & \underline{72.7} \\
\rowcolor{lightblue}
GPT-5-20250807~\cite{2022chatgpt} & N/A & 72.9 & 81.9 & \textbf{72.0} & \textbf{81.2} & \textbf{60.9} & 71.1 & 70.0 \\
\rowcolor{lightblue}
Gemini-2.5-Pro~\cite{team2023gemini} & N/A & 72.5 & 80.9 & 69.1 & 76.9 & 56.3 & \textbf{78.0} & \textbf{73.8} \\
\rowcolor{lightblue}
GPT-5-mini-20250807~\cite{2022chatgpt} & N/A & 70.1 & 79.2 & \underline{69.9} & 74.4 & \underline{56.5} & 70.8 & 69.8 \\
RBDash-v2.0-Thinking & 34B & 64.8 & \underline{82.2} & 59.9 & 65.9 & 46.9 & 67.5 & 66.7 \\
\rowcolor{lightblue}
Doubao-1.5-pro~\cite{Doubao} & N/A & 61.6 & 78.6 & 51.5 & 64.7 & 44.9 & 65.7 & 64.2 \\
Logics-Thinking-32B & 34B & 60.8 & 80.4 & 51.4 & 70.3 & 42.1 & 63.0 & 57.9 \\
\rowcolor{lightblue}
GPT-5-nano-20250807~\cite{2022chatgpt} & N/A & 60.7 & 73.1 & 59.7 & 66.6 & 47.9 & 59.4 & 57.5 \\
R-4B & 4.8B & 57.6 & 77.8 & 48.8 & 61.0 & 42.9 & 54.7 & 60.2 \\
\rowcolor{lightblue}
360zhinao-pro-vision & N/A & 56.8 & 80.9 & 47.9 & 44.8 & 44.5 & 58.8 & 64.0 \\
\shline
\end{tabular}}  
\vspace{0.5mm}
\caption{The evaluation results of LMMs on image reasoning benchmarks.
The table displays the top-10 LMMs, including both commercial APIs and open-source LMMs(\textbf{till 2025.11.21}), in the descending order of average score.
}
\label{tab:reasoning_main_results}
\vspace{-4mm}
\end{table*}

%% file: sections/evaluation.tex
\section{Evaluation Results}
\subsection{General VQA Benchmarks}
Utilizing VLMEvalKit, one can conduct comprehensive evaluations of an LMM 
across numerous benchmarks to gain a thorough understanding of its strengths and weaknesses. 
We publish all evaluation results on  \href{https://huggingface.co/spaces/opencompass/open_vlm_leaderboard}{OpenVLM Leaderboard}. 
Our core leaderboard is based on evaluations from eight distinct benchmarks:
1. MMBench v1.1 [test]~\cite{liu2023mmbench}\footnote{We report the average score of English and Chinese test splits.} (all-round capability);
2. MMStar~\cite{chen2024we} (data contamination); 
3. MMMU~\cite{yue2023mmmu} [val] (multi-modal examination);
4. MathVista [mini-test]~\cite{lu2023mathvista} (multi-modal math);
5. HallusionBench~\cite{liu2023hallusionbench} (hallucination \& illusion);
6. AI2D [test]~\cite{kembhavi2016diagram} (diagram understanding); 
7. OCRBench~\cite{liu2023hidden} (text understanding); 
8. MMVet~\cite{yu2023mm} (subjective evaluation). 
The selection encompasses a diverse array of tasks, 
and the average score across these benchmarks serves as a reliable indicator 
of the general capabilities of LMMs. 
In \cref{tab:vqa_main_results}, we present the performance of 
top-10 commercial APIs and the top-10 open-source models evaluated on this suite of benchmarks.

\input{sections/video_main_table}

\subsection{Image Reasoning Benchmarks}
With the rapid advancement of LMMs, reasoning capabilities have garnered increasing attention. 
To systematically assess these capabilities, we have compiled several widely recognized multi-modal benchmarks designed for reasoning tasks, including MathVista~\cite{lu2023mathvista}, MathVision~\cite{wang2025measuring}, MathVerse\cite{zhang2024mathverse}, DynaMath\cite{zou2024dynamath}, WeMath\cite{qiao2024we}, and LogicVista\cite{xiao2024logicvista}. 
Evaluations were conducted on \href{https://rank.opencompass.org.cn/leaderboard-multimodal-reasoning}{OpenVLM  Reasoning Leaderboard}. 
The selected benchmarks primarily emphasize mathematical and puzzle-solving problems, spanning areas such as geometry, algebra, logic, and related domains. The performance metrics of the top-10 models are summarized in \cref{tab:reasoning_main_results}.

Cut down to 2025.11, in multi-modal reasoning tasks, a significant performance gap remains evident between commercial APIs and open-source LMMs. Generally, commercial APIs demonstrate a marked advantage over their open-source counterparts. Notably, the top four performing LMMs are all proprietary APIs. RBDash-v2.0-Thinking, the highest-performing open-source LMM on reasoning benchmarks, achieves an average score of 64.8, which still highlights a considerable gap compared to leading models like Seed1.5-VL~\cite{guo2025seed15vl} (73.3) and GPT-5 (72.9). 
Further analysis reveals that most high-performing models achieve accuracy exceeding 80\% on MathVista~\cite{lu2023mathvista}. In contrast, only a few leading models (such as GPT-5, Seed1.5-VL, and Gemini-2.5-Pro) achieve accuracy greater than 50\% on DynaMath~\cite{zou2024dynamath}, underscoring the severe challenges posed by multi-step mathematical reasoning.

\subsection{Video Benchmarks}

Video has become essential in daily life, driving communication, learning, and entertainment. With the rapid rise of video content and progress in LMM capability, video understanding ability is increasingly vital for LMMs. To fully examine the video understanding capability, we have added support for some popular common video understanding benchmarks, including MMBench-Video~\cite{fang2024mmbench}, Video-MME~\cite{fu2024videomme}, MLVU~\cite{zhou2024mlvu}, TempCompass~\cite{liu2024tempcompass}, MVBench~\cite{li2024mvbench}, and other video datasets. Besides, in order to have a glimpse into the overall video understanding performance of LMMs, we also selected some representative video benchmarks and evaluated over forty LMMs, with the results presented on the \href{https://huggingface.co/spaces/opencompass/openvlm_video_leaderboard}{OpenVLM Video Leaderboard}. The selected
benchmarks primarily cover various question types (multiple choice, open-ended questions, true/false questions, etc.) and various video lengths, mainly examining the model's general video understanding ability. The performance metrics of the top-10 models are summarized in \cref{tab:video_main_results}.

Due to the limited resources of commercial APIs and the significant costs associated with video understanding evaluations, we focused on testing the video understanding capabilities of open-source LMMs. It's easy to observe that InternVL and Qwen Series demonstrate remarkable performance in video understanding benchmarks. InternVL3-78B~\cite{zhu2025internvl3} achieves the highest average score of 72.7, and Qwen2.5 lags slightly behind them, mainly revealed in Video-MME~\cite{fu2024videomme} and MLVU~\cite{zhou2024mlvu}. With a 64-frame input window, Gemini-2.0-flash~\cite{team2023gemini} still lags behind the state of the art on several tasks, possibly because it refuses to answer some questions. GPT-4o~\cite{2022chatgpt} achieves 63.1 with only 16 frames, showing that commercial APIs can still compete in video understanding even under a short context.

\subsection{Closed-Source Evaluation}
\input{sections/closed_source_main_table}
As multi-modal large models evolve from basic perceptual understanding towards higher-order intelligence encompassing cognition, reasoning, and creation, evaluating them on more challenging, uncompromised datasets becomes critical. To thoroughly assess these advanced capabilities and prevent data contamination, we introduce a comprehensive evaluation based on closed-source benchmarks. The benchmark evaluates models across diverse dimensions, including General Perception, Spatial Understanding, Infographic Understanding, MultiModal Reasoning, and MultiModal Creation. The results of the top-10 models are presented in \cref{tab:closed_source_results}.

\noindent\textbf{Overall Performance.} Proprietary commercial APIs continue to push the boundaries of multi-modal intelligence, with models like Gemini-3.1-Pro-Preview (\textbf{66.62}) and Doubao-Seed-2.0-Pro (\textbf{63.19}) securing top positions. However, the open-source community is rapidly closing the gap. Notably, Qwen3.5-397B (\textbf{65.41}) and Kimi-k2.5 (\textbf{63.05}) demonstrate exceptional performance, standing shoulder-to-shoulder with leading closed-source APIs.

\noindent\textbf{General Perception \& Infographics:} Leading models perform well in general perception, with Gemini-3.1-Pro-Preview and Qwen3.5-397B scoring above 73. Similarly, infographic understanding is dominated by top-tier models, highlighting their robust capability in extracting structured information.

\noindent\textbf{MultiModal Reasoning:} This dimension remains a significant differentiator. Gemini-3.1-Pro-Preview leads with a score of 66.25, while the majority of models struggle to surpass the 60-point threshold, indicating that multi-step logical reasoning on complex visual inputs is still in an iterative improvement phase.

\noindent\textbf{MultiModal Creation:} Both open-source and proprietary models excel in creation tasks (e.g., generating travel journals or creative narratives). Claude-Opus-4-6 (91.97) and Kimi-k2.5 (90.05) lead this dimension, demonstrating human-like contextual understanding and expressive generation.

\noindent\textbf{Spatial Understanding Bottleneck:} Despite advancements in other areas, spatial understanding remains a universal bottleneck across all models. The highest score achieved in this dimension is only 42.0 (Doubao-Seed-2.0-Pro and Kimi-k2.5), revealing a critical area for future breakthrough in understanding 3D spatial relationships and physical environments.

% We also established the  OpenVLM Video Leaderboard . The selected
% benchmarks primarily emphasize mathematical and puzzle-
% solving problems, spanning areas such as geometry, alge-
% bra, logic, and related domains. The performance metrics
% of the top-10 models are summarized in Tab. 3

%% file: sections/video_main_table.tex
\begin{table*}[t]
\centering
\resizebox{\linewidth}{!}{
\tablestyle{6pt}{1.5}
\begin{tabular}{ccccccccc}
\shline
\rowcolor{lightgray}
Method            & Param.   & Frame. & Avg. Score & MVBench & Video-MME(w/o subs) & MMBench-Video & TempCompass & MLVU  \\
\shline
InternVL3-78B~\cite{zhu2025internvl3}  & 78.4     & 64 & \textbf{72.7} & \textbf{79.2} & \textbf{73.1} & 1.81 & 77.0 & \textbf{74.0}  \\
InternVL3-38B~\cite{zhu2025internvl3}  & 38.4     & 64 & \underline{71.9} & \underline{76.0} & \underline{72.1} & 1.80 & \underline{78.5} & \underline{72.7}  \\
Qwen2.5-VL-72B~\cite{Qwen2.5-VL}  & 73.4     & 64 & 68.6 & 71.3 & 68.6 & 1.84 & 77.7 & 64.1  \\
InternVL2.5-78B~\cite{chen2024expanding}  & 78.4     & 64 & 68.1 & 75.6 & \underline{72.1} & \underline{1.98} & 56.0 & 70.9  \\
Qwen2-VL-72B~\cite{wang2024qwen2}  & 73.4     & 64 & 67.7 & 70.2 & 67.3 & 1.70 & \textbf{79.4} & 65.2  \\
InternVL3-8B~\cite{zhu2025internvl3}  & 7.94     & 64 & 66.8 & 73.2 & 66.0 & 1.69 & 70.4 & 68.0  \\
LLaVA-Video-72B-Qwen2~\cite{zhang2024llavavideo} & 7.94     & 64 & 66.2 & 63.1 & 70.5 & 1.71 & 70.4 & 70.0  \\
Aria~\cite{aria} & 25.3     & 64 & 66.2 & 67.9 & 66.0 & 1.81 & 69.6 & 67.1  \\
\rowcolor{lightblue}
Gemini-2.0-flash~\cite{team2023gemini}  & N/A        & 64 & 64.6 & 59.4 & 71.0 & \textbf{2.01} & 57.9 & 68.0\\
\rowcolor{lightblue}
GPT-4o-20240806~\cite{2022chatgpt}  & N/A        & 16 & 63.1 & 57.5 & 67.9 & 1.87 & 72.7 & 54.9\\
Qwen2.5-Omni-7B~\cite{xu2025qwen25omni}  & 10.7        & 64 & 64.5 & 69.0 & 64.1 & 1.65 & 70.7 & 63.6\\
\shline
\end{tabular}}

\vspace{0.5mm}
\caption{The evaluation results of LMMs on video understanding benchmarks. 
The table displays the top-10 LMMs, including both commercial APIs and open-source LMMs(\textbf{till 2025.06.25}), in the descending order of average score. Due to resource limitations, we only tested a few commercial APIs.
}
\label{tab:video_main_results}
\vspace{-4mm}
\end{table*}

%% file: sections/closed_source_main_table.tex
\begin{table*}[t]
\centering
\resizebox{\linewidth}{!}{
\tablestyle{4pt}{1.5}
\begin{tabular}{lcccccccc}
\shline
\rowcolor{lightgray}
Method & Param. & Avg. Score & Gen. Percept. & Spatial Und. & Infographic Und. & MM Reason. & MM Creation \\
\shline
\rowcolor{lightblue}
Gemini-3.1-Pro-Preview & N/A & \textbf{66.62} & \textbf{74.0} & 38.5 & \textbf{76.0} & \textbf{66.25} & 78.73 \\
Qwen3.5-397B & 397B & \underline{65.41} & \underline{73.0} & \underline{41.5} & \underline{75.0} & \underline{63.75} & 75.44 \\
\rowcolor{lightblue}
Doubao-Seed-2.0-Pro & N/A & 63.19 & 70.0 & \textbf{42.0} & 68.0 & 58.13 & 82.90 \\
Kimi-k2.5 & 1T & 63.05 & 63.0 & \textbf{42.0} & 67.0 & 58.13 & \underline{90.05} \\
\rowcolor{lightblue}
SenseNova-V6-5-Pro & N/A & 55.61 & 49.0 & 34.5 & 58.0 & 51.25 & 89.67 \\
\rowcolor{lightblue}
Claude-Opus-4-6 & N/A & 55.16 & 42.0 & 36.5 & 58.0 & 51.25 & \textbf{91.97} \\
GLM4\_6V & 106B & 52.54 & 50.0 & 33.5 & 59.0 & 45.62 & 81.50 \\
\rowcolor{lightblue}
GPT-5.4 & N/A & 51.55 & 46.0 & 31.5 & 58.0 & 45.62 & 82.57 \\
Step3-VL-10B & 10B & 48.80 & 38.0 & 34.0 & 41.0 & 51.88 & 76.07 \\
Ovis2.6-30B-A3B & 30B & 48.37 & 44.0 & 26.5 & 49.0 & 54.37 & 62.00 \\
\shline
\end{tabular}}  
\vspace{0.5mm}
\caption{The evaluation results of LMMs on closed-source benchmarks.
The table displays the top-10 LMMs, including both commercial APIs and open-source LMMs(\textbf{till 2026.02}), in the descending order of average score.
We assess five core capabilities: General Perception, Spatial Understanding, Infographic Understanding, MultiModal Reasoning, and MultiModal Creation.}
\label{tab:closed_source_results}
\vspace{-4mm}
\end{table*}

%% file: sections/discussion.tex
\section{Discussion}

We have released VLMEvalKit, 
an open-source toolkit designed for the evaluation of large multi-modality models. 
VLMEvalKit encompasses a comprehensive collection of 
over 300 LMMs and more than 100 multi-modal benchmarks.
The codebase is structured with simplicity in mind, 
facilitating the easy integration of new LMMs or benchmarks. 
Our framework is thoughtfully designed to extend its capabilities beyond the vision modality and incorporate other modalities like audio.
Moving forward, the development of VLMEvalKit will focus on 
expanding the repertoire of LMMs and benchmarks for video and other modalities. 
We are optimistic that this repository, along with all released resources, 
will contribute to advancing research in multi-modal learning.

%% file: sections/appendix.tex
\appendix
\section{Benchmarks}
In this section, we list all supported benchmarks with a brief introduction and categorize them into multiple categories. 

\subsection{General Capability}
\begin{itemize}[itemsep=1pt,topsep=0pt,parsep=0pt]
    \item \textbf{MMBench Series}~\cite{liu2023mmbench}: 
    The MMBench Series, which includes MMBench and its Chinese version MMBench-CN, serves as a pioneering benchmark for assessing Large Vision-Language Models (LVLMs) across 20 distinct capabilities. MMBench-CN adapts the questions and choices of MMBench into Chinese based on GPT-4. These benchmarks are designed to promote a more precise and comprehensive evaluation of VLMs.
    \item \textbf{SEEDBench \& SEEDBench2}~\cite{li2023seed}: SEEDBench comprises 19,000 multiple-choice questions that span 12 dimensions, while SEEDBench2 expands this dataset to 24,000 questions across 27 dimensions. Both benchmarks provide a robust framework for evaluating the multifaceted capabilities of AI models.
    \item \textbf{MME}~\cite{Fu2023MMEAC}: a holistic benchmark aimed at evaluating MLLMs. It assesses perception and cognition across 14 subtasks, featuring manually curated instruction-answer pairs to prevent data leakage and ensure a fair comparison without the need for prompt engineering.
    \item \textbf{MMT-Bench}~\cite{ying2024mmt}: a comprehensive benchmark designed to assess LVLMs across massive multimodal tasks requiring expert knowledge and deliberate visual recognition, localization, reasoning, and planning.
    \item \textbf{MME-RealWorld}~\cite{zhang2024mme}: a large-scale benchmark for evaluating MLLMs in real-world scenarios. It comprises 29,429 question-answer pairs and encompasses 43 subtasks, providing a rich dataset for testing the practical applications of MLLMs. 
\end{itemize}

\subsection{Text Recognition \& Understanding}
\begin{itemize}[itemsep=1pt,topsep=0pt,parsep=0pt]
    \item \textbf{TextVQA}~\cite{textvqa}: a dataset for assessing visual reasoning based on textual content within images. TextVQA challenges models to read and reason about the text present in images to answer corresponding questions. To accomplish this, models must integrate the textual information found in the images and utilize it to formulate responses to TextVQA's queries.
    \item \textbf{OCRVQA}~\cite{mishra2019ocr}: a benchmark for visual question answering by reading text in images. It introduces the OCR-VQA-200K dataset, which consists of over 200,000 book cover images. This study integrates techniques from Optical Character Recognition (OCR) and Visual Question Answering (VQA) to tackle the novel challenges associated with this dataset.
    \item \textbf{OCRBench}~\cite{liu2023hidden}: a comprehensive evaluation benchmark designed to assess the Optical Character Recognition (OCR) capabilities of LMMs across various text-related visual tasks, including text recognition, scene text-centric VQA, document-oriented VQA, key information extraction, and handwritten mathematical expression recognition.
    \item \textbf{DocVQA}~\cite{mathew2021docvqa}: designed to encourage a “purpose-driven” point of view in Document Analysis and Recognition research, where the document content is extracted and used to respond to high-level tasks defined by the human consumers of this information.
    \item \textbf{olmOCRBench}~\cite{poznanski2025olmocr}: a recent OCR benchmark accompanying the olmOCR document understanding pipeline, focusing on PDF-style scientific and historical documents with strict layout-aware evaluation.
    \item \textbf{OceanOCRBench}~\cite{chen2025oceanocr}: a domain-oriented OCR benchmark covering complex Chinese-English bilingual scenes and structured business documents.
    % \item \textbf{SciDocBench}: a scientific document benchmark that decouples reasoning evaluation from answer scoring, allowing fine-grained probing of layout, table, and figure-centric reasoning.}
\end{itemize}

\subsection{Structuralized Content Understanding}
\begin{itemize}[itemsep=1pt,topsep=0pt,parsep=0pt]
    \item \textbf{ChartQA}~\cite{masry2022chartqa}: a benchmark designed to assess question answering skills related to charts, with an emphasis on complex visual and logical reasoning. The dataset comprises 9,600 human-written questions and 23,100 machine-generated questions based on chart summaries. It challenges models to perform intricate reasoning that includes logical and arithmetic operations as well as to analyze visual chart features.
    \item \textbf{InfoVQA}~\cite{mathew2022infographicvqa}: a dataset aimed at enhancing the automatic understanding of infographic images through VQA. It includes a diverse collection of infographics paired with natural language questions and answers, focusing on the need for models to reason across document layouts, textual content, graphical elements, and data visualizations.
    \item \textbf{TableVQA-Bench}~\cite{kim2024tablevqa}: a benchmark created for visual question answering focused on table data. It integrates images and question-answer pairs that were not previously available in existing datasets. The benchmark consists of 1,500 QA pairs generated from text-formatted tables using a large language model (LLM).
    \item \textbf{SEEDBench2-Plus}~\cite{li2024seed}: a benchmark that includes 2,300 visual questions covering three categories (Charts, Maps, and Webs) and 63 fine-grained data types found in real-world scenarios. It is designed to evaluate the capabilities of multimodal large language models in understanding text-rich visual content.
    \item \textbf{ChartMuseum}~\cite{tang2026chartmuseum}: a chart reasoning benchmark stress-testing the visual reasoning capabilities of large vision-language models on diverse chart types and complex multi-step questions.
    \item \textbf{ChartQAPro}~\cite{masry2025chartqapro}: an extended, more diverse, and more challenging successor to ChartQA, covering a wider distribution of chart families and a stricter evaluation protocol.
    \item \textbf{ChartX}~\cite{xia2025chartx}: a versatile chart benchmark accompanied by the ChartVLM foundation model, covering a wide spectrum of chart types and reasoning tasks.
\end{itemize}
 
\subsection{Subjective (LLM Judge)}
\begin{itemize}[itemsep=1pt,topsep=0pt,parsep=0pt]
    \item \textbf{MM-Vet}~\cite{yu2023mm}: an evaluation benchmark for LMMs designed to assess their performance on complex multimodal tasks. It provides a systematic structure for these tasks and proposes unified evaluation metrics. The benchmark utilizes Large Language Model (LLM)-based evaluators to assess a variety of question and answer types, ensuring a comprehensive analysis of LMM capabilities.
    \item \textbf{LLaVABench}~\cite{liu2023visual}: a dataset created to evaluate the capabilities of LMMs in more challenging tasks and their ability to generalize to new domains. It consists of a diverse set of 24 images with 60 questions in total, including indoor and outdoor scenes, memes, paintings, sketches, etc., and each image with a highly-detailed and manually-curated description and a proper selection of questions.
\end{itemize}

\subsection{Mathematics \& Examination}
\begin{itemize}[itemsep=1pt,topsep=0pt,parsep=0pt]
    \item \textbf{MathVista}~\cite{lu2023mathvista}: a benchmark designed to evaluate the mathematical reasoning capabilities of LLMs and LMMs in visual contexts. It comprises 6,141 examples from 28 existing multimodal datasets and three new datasets (IQTest, FunctionQA, and PaperQA), challenging models with deep visual understanding and compositional reasoning.
    \item \textbf{MathVision}~\cite{lu2023mathvista}: a dataset addressing limitations in existing benchmarks for evaluating mathematical reasoning in visual contexts. It includes 3,040 high-quality mathematical problems sourced from actual competitions, spanning 16 disciplines and varying in difficulty across five levels.
    \item \textbf{ScienceQA-IMG}~\cite{lu2022learn}: a multimodal benchmark for answering science-related questions. It contains approximately 21,000 questions and annotations, promoting reasoning through the Chain of Thought (CoT) approach.
    \item \textbf{MMMU}~\cite{yue2023mmmu}: a comprehensive benchmark for evaluating multimodal models on tasks that require expert-level knowledge. It features 11,500 questions across six disciplines and 183 subfields.
    % \item \textbf{AI2D}~\cite{kembhavi2016diagram}: a dataset of over 5000 grade school science diagrams with annotations and multiple choice questions.    
\end{itemize}

\subsection{Low-level \& Aesthetics}
\begin{itemize}[itemsep=1pt,topsep=0pt,parsep=0pt]
    \item \textbf{AesBench}~\cite{huang2024aesbench}: a benchmark for evaluating aesthetic perception capabilities of MLLMs with an expert-labeled database and from four shallow-to-deep perspectives.
    \item \textbf{Q-Bench}~\cite{wu2023q}: a benchmark for evaluating MLLMs in low-level vision tasks, including perception, description, and assessment. Only multi-choice questions are included in our implementation. 
    \item \textbf{A-Bench}~\cite{zhang2024bench}: a benchmark for evaluating LMMs in assessing AI-generated images, including 2864 AIGIs from 16 text-to-images models.
\end{itemize}

\subsection{Long-Context, Multi-Image \& Multi-Turn}
\begin{itemize}[itemsep=1pt,topsep=0pt,parsep=0pt]
    \item \textbf{DUDE}~\cite{van2023document}: a comprehensive benchmark aimed at advancing Document AI by addressing limitations in current methods for visually-rich documents. The benchmarks include 6315 visual questions, with an average of 5.7 images per question. 
    \item \textbf{MMLongBench-Doc}~\cite{ma2024mmlongbench}: a benchmark designed to evaluate understanding of long-context documents in LVLMs. It consists of 1,091 expert-annotated questions based on 135 lengthy PDF documents, averaging 47.5 pages and 21,214 tokens each. This benchmark uniquely requires that answers be derived from various multimodal sources, including text, images, charts, tables, and layout structures, and 33\% of the questions require cross-page evidence.
    \item \textbf{SlideVQA}~\cite{tanaka2023slidevqa}: a multi-image document VQA dataset, consisting of 52k+ slide images from 2.6k+ slide decks and 14.5k questions. It requires complex reasoning, including numerical and multihop reasoning, and provides annotated arithmetic expressions for numerical answers.
    \item \textbf{BLINK}~\cite{fu2024blink}: a benchmark with 14 challenging visual perception tasks for multimodal LLMs, with around 60\% of visual questions consisting of multiple images as inputs.
    \item \textbf{MuirBench}~\cite{wang2024muirbench}: a comprehensive benchmark for multimodal LLMs on multi-image understanding with 12 tasks, 10 relation categories, 11264 images and 2600 questions.
    \item \textbf{MMDU}~\cite{liu2024mmdu}: a benchmark designed to evaluate VLMs capabilities in multi-turn, multi-image dialogues, addressing their limitations in handling real-world, complex conversational scenarios. MMDU features longer dialogues with up to 27 turns and 20 images, challenging current LVLMs.
\end{itemize}

\subsection{Hallucination}
\begin{itemize}[itemsep=1pt,topsep=0pt,parsep=0pt]
    \item \textbf{HallusionBench}~\cite{liu2023hallusionbench}: a diagnostic benchmark for evaluating large visual-language models' reasoning about image-context relationships, focusing on visual illusions and language hallucinations. Consists of 346 images and 1,129 questions to assess response consistency, failure modes, and logical reasoning.
    \item \textbf{POPE}~\cite{li2023evaluating}: a benchmark for evaluating object hallucination with three tracks (random, popular, and adversarial) and a total of $\sim$ 9k cases.
\end{itemize}

\subsection{Multilingual}
\begin{itemize}[itemsep=1pt,topsep=0pt,parsep=0pt]
    \item \textbf{MMMB and Multilingual MMBench}~\cite{sun2024parrot}: MMMB is a multilingual multimodal benchmark with six languages, 15 categories, and 12k questions. Multilingual MMBench extends MMBench-DEV to six languages.
    \item \textbf{MT-VQA}~\cite{tang2024mtvqa}: a benchmark designed for multilingual Text-Centric Visual Question Answering, addressing limitations of previous datasets that focus on high-resource languages and rely on translation-based approaches. The work introduces high-quality human annotations across nine diverse languages, avoiding "visual-textual misalignment" and improving multilingual scene understanding.
    \item \textbf{Aya Vision Bench}~\cite{dash2025aya}: a multilingual vision-language benchmark designed to evaluate cross-lingual visual question answering across a diverse set of languages.
\end{itemize}

\subsection{Miscellaneous}
\begin{itemize}[itemsep=1pt,topsep=0pt,parsep=0pt]
    \item \textbf{MLLMGuard (Safety)}~\cite{gu2024mllmguard}: a multi-dimensional safety evaluation suite for MLLMs, including a bilingual image-text evaluation dataset, inference utilities, and a set of lightweight evaluators.
    \item \textbf{GMAI-MMBench (Medical)}~\cite{chen2024gmai}: a comprehensive multimodal evaluation benchmark for VLMs in the medical field across multiple datasets, image modalities, and clinical tasks.
    \item \textbf{RealWorldQA (Autonomous Driving)}~\cite{RealWorldQA}: a benchmark for evaluating real-world spatial understanding of multimodal AI models with over 700 images from various scenarios.
    \item \textbf{COCO Caption (Captioning)}~\cite{lin2014microsoft}: Contains 5,000 samples from the COCO Caption Validation set and prompts VLMs to describe the images.
    \item \textbf{MMStar (Data Contamination)}~\cite{chen2024we}: a vision-indispensable multi-modal benchmark addressing issues of unnecessary visual content and data leakage in evaluating LVLMs. Comprises 1,500 curated samples and proposes new metrics for data leakage.
    \item \textbf{TaskMeAnything ImageQA}~\cite{zhang2024task}: a benchmark generation engine for creating tailored benchmarks with an extendable taxonomy of visual assets.
    \item \textbf{A-OKVQA}~\cite{schwenk2022okvqa}: a benchmark for assessing VQA using world knowledge and commonsense reasoning with around 25k crowdsourced questions.
    \item \textbf{VCR-wiki}~\cite{zhang2024vcr}: Visual Caption Restoration (VCR) is a new vision-language task that requires models to restore partially obscured texts in images using pixel-level hints. Unlike traditional tasks that often rely on optical character recognition or masked language modeling, VCR emphasizes the integration of visual, textual, and contextual cues to achieve accurate text restoration.
\end{itemize}

\subsection{Video Understanding}
\begin{itemize}[itemsep=1pt,topsep=0pt,parsep=0pt]
\item \textbf{MMBench-Video}~\cite{fang2024mmbench}: a long-form, multi-shot benchmark with about 600 YouTube videos (30 s–6 min) drawn from 16 everyday domains. It offers roughly 2000 volunteer-written Q\&A pairs that probe 26 fine-grained skills and uses a GPT-4 adjudication pipeline for reliable scoring, making it a concise testbed for holistic video understanding.

\item \textbf{Video-MME}~\cite{fu2024videomme}: the first “full-spectrum” evaluation for video LLMs, containing 900 clips plus 2700 human-annotated Q\&A pairs. Videos span short to hour-long segments across six visual domains, and the benchmark explicitly mixes frames, audio, and subtitles to examine cross-modal reasoning over varied temporal scales. 

\item \textbf{MVBench}~\cite{li2024mvbench}: targets temporal reasoning by converting 20 classic vision tasks into dynamic versions that cannot be solved with a single frame. Automatic conversion of public annotations into multiple-choice QAs ensures fairness, and early results show mainstream MLLMs still fall short on many perception-to-cognition temporal skills.

\item \textbf{MLVU}~\cite{zhou2024mlvu}: focuses on long-video comprehension, gathering clips from 3 min up to 2 h and organising nine tasks that demand both global narrative understanding and fine-grained detail tracking. Initial evaluations (20 models, incl. GPT-4o) reveal that even the best system scores only 64.6\%, underscoring the difficulty of sustained reasoning over extended contexts.

\item \textbf{TempCompass}~\cite{liu2024tempcompass}: an benchmark designed to gauge temporal perception. It covers diverse temporal aspects (action granularity, motion speed, event order, attribute change, etc.) and multiple task formats (multi-choice, yes/no, caption matching/generation). Carefully crafted “conflicting videos” reduce single-frame shortcuts, offering a stricter measure of whether models truly exploit temporal cues. 

\item \textbf{Video-MME-v2}~\cite{fu2026videommev2}: the next-generation full-spectrum video understanding benchmark succeeding Video-MME, with broadened domain coverage and stricter cross-modal evaluation.
\item \textbf{VideoMMMU}~\cite{hu2026videommmu}: a video-centric extension of MMMU, evaluating expert-level multi-disciplinary knowledge over instructional videos.
\item \textbf{VSI-Bench}~\cite{yang2025vsibench}: a benchmark probing visual-spatial intelligence in video understanding, including object counting, relative orientation, and ego-motion reasoning.
\end{itemize}

\subsection{Spatial Reasoning}
\begin{itemize}[itemsep=1pt,topsep=0pt,parsep=0pt]
    \item \textbf{TopViewRS}~\cite{li2024topviewrs}: evaluates vision-language models as top-view spatial reasoners over indoor scenes.
    \item \textbf{ERQA}~\cite{team2025geminierqa}: an embodied reasoning question-answering benchmark covering grounded visual reasoning in real-world scenes.
    \item \textbf{RefSpatialBench}~\cite{zhou2026refspatial}: a spatial reference understanding benchmark requiring 3D-aware grounding of natural-language expressions.
\end{itemize}

\subsection{GUI Grounding \& Understanding}
\begin{itemize}[itemsep=1pt,topsep=0pt,parsep=0pt]
    \item \textbf{OSWorld-G}~\cite{xie2026osworldg}: a GUI grounding benchmark derived from the OSWorld environment, focused on element-level localization on real desktop screenshots.
    \item \textbf{VenusBench-GD}~\cite{team2026venusbench}: a multi-platform GUI grounding benchmark covering web, mobile, and desktop UIs.
\end{itemize}

\subsection{Safety, Bias \& Alignment}
\begin{itemize}[itemsep=1pt,topsep=0pt,parsep=0pt]
    \item \textbf{VLMsAreBiased}~\cite{vo2025vlmsarebiased}: a benchmark evaluates how strong prior knowledge of popular subjects severely impairs a vision language model's (VLM) ability to perform objective visual tasks.
    \item \textbf{MMSafetyBench}~\cite{liu2024mmsafetybench}: a multimodal safety benchmark covering jailbreak, toxic content, and unsafe instruction scenarios.
    \item \textbf{Flames}~\cite{huang2024flames}: a value-alignment benchmark for LLMs/VLMs in Chinese, focused on harmlessness and ethical robustness.
    % \item \textbf{MSSBench}: a multimodal situational safety benchmark unveiling visual leakage and unsafe outputs.
    % \item \textbf{SIUO}: ``Safe Inputs but Unsafe Outputs'' --- a benchmark for cross-modality safety alignment.
    % \item \textbf{M3oralBench}: a multimodal moral evaluation benchmark for large vision-language models.
    % \item \textbf{XSTest}: an over-refusal calibration benchmark testing whether models reject benign prompts that surface-resemble unsafe ones.
\end{itemize}

\subsection{Medical, Scientific \& Domain Reasoning}
\begin{itemize}[itemsep=1pt,topsep=0pt,parsep=0pt]
    \item \textbf{MedQ-Bench}~\cite{liu2025medqbench}: a medical image quality assessment benchmark for multimodal LLMs.
    \item \textbf{HiPhO}~\cite{yu2025hipho}: a high-school-physics-olympiad benchmark stress-testing multi-step reasoning of (M)LLMs.
    \item \textbf{MMSci}~\cite{li2024mmsci}: a multimodal benchmark spanning 72 scientific disciplines (primarily in the natural sciences), specifically designed to evaluate the comprehension and reasoning capabilities of AI models on PhD-level scientific literature.
\end{itemize}

\section{Acknowledgements}

We make our best effort to list all known contributions to the repository. If some of your contributions are missing, please contact opencompass@pjlab.org.cn.

\subsection{Contributors with 3+ Major Contributions}
We list contributors who have made 3+ significant contributions to the development of VLMEvalKit.

\textbf{Qualified Contributors (2025.02):}
\begin{itemize}[itemsep=0pt,topsep=0pt,parsep=0pt]
    \item \href{https://github.com/PhoenixZ810}{PhoenixZ810} (Xiangyu Zhao): The contributor helped support WeMath~\cite{qiao2024we}, LogicVista~\cite{xiao2024logicvista}, MM-AlignBench~\cite{zhao2025omnialign}, Video-ChatGPT~\cite{maaz2023video}, Chat-UniVI~\cite{jin2024chat}, and Llama-VID~\cite{li2024llama}.
    \item \href{https://github.com/amitbcp}{amitbcp} (Amit Agarwal): The contributor helped support MUIRBench~\cite{wang2024muirbench}, Phi-3.5~\cite{abdin2024phi}, Idefics3~\cite{laurenccon2024building}, VILA~\cite{lin2024vila}, xGen-MM~\cite{xue2024xgen}, and MVTamperBench~\cite{agarwal2024mvtamperbench}.
    \item \href{https://github.com/czczup}{czczup} (Zhe Chen): The contributor helped support the InternVL Series~\cite{chen2023internvl,chen2024far,gao2024mini} (V1.5, Mini-InternVL, V2, etc.).
    \item \href{https://github.com/Mor-Li}{Mor-Li} (Mo Li): The contributor helped support LLaVA-OneVision~\cite{li2024llama}, GQA~\cite{hudson2019gqa}, and developed the readthedocs site for VLMEvalKit.
    \item \href{https://github.com/mayubo2333}{mayubo2333} (Yubo Ma): The contributor helped support MMLongBench~\cite{ma2024mmlongbench}, SlideVQA~\cite{tanaka2023slidevqa}, and DUDE~\cite{van2023document}.
    \item \href{https://github.com/sun-hailong}{sun-hailong} (Hailong Sun): The contributor helped support A-OKVQA~\cite{schwenk2022okvqa}, Parrot, MMMB, and MTL-MMBench~\cite{sun2024parrot}.
    \item \href{https://github.com/Cuiunbo}{Cuiunbo} (Junbo Cui): The contributor helped support OmniLMM-12B, MiniCPM-V Series~\cite{yao2024minicpm} (V1, V2, V2.5).
    \item \href{https://github.com/yfzhang114}{yfzhang114} (yifan zhang) The contributor helped support Slime~\cite{zhang2024beyond}, MME-RealWorld~\cite{zhang2024mme}, and Amber~\cite{wang2023amber}.
    \item \href{https://github.com/runninglsy}{runninglsy} (Shiyin Lu) The contributor helped support Ovis Series~\cite{lu2024ovis}.
    \item \href{https://github.com/tackhwa}{tackhwa} (Tack Hwa Wong) The contributor helped support Eagle X~\cite{shi2024eagle}, Moondream, Kosmos2~\cite{peng2023kosmos}.
    \item \href{https://github.com/Weiyun1025}{Weiyun1025} (Weiyun Wang) The contributor helped support InternVL-COT evaluation, InternVL2-8B-MPO, InternVL2.5-MPO~\cite{wang2024enhancing}.
    \item \href{https://github.com/Myhs-phz}{Myhs-phz} (Peiheng Zhou) The contributor helped support MIA-Bench~\cite{qian2024mia}, VizWiz~\cite{vizwiz}, OlympiadBench~\cite{he2024olympiadbench}, CMMMU~\cite{zhang2024cmmmu}.
    \item \href{https://github.com/BradyFU}{BradyFU} (Chaoyou Fu) The contributor helped support Video-MME~\cite{fu2024video}, Vita 1.0~\cite{fu2024vita}, Vita 1.5~\cite{fu2025vita}.
    \item \href{https://github.com/OliverLeeXZ}{OliverLeeXZ} (Xiaozhe Li) The contributor helped support Emu3-[Chat/Gen]~\cite{wang2024emu3}, Moonshot APIs, Grok APIs.
\end{itemize}

\textbf{Qualified Contributors (2025.06):}
\begin{itemize}[itemsep=0pt,topsep=0pt,parsep=0pt]
    \item \href{https://github.com/chenjix}{chenjix} (JiXuan Chen): The contributor helped support Humanity's last exam~\cite{phan2025hle}, LLama4~\cite{touvron2023llama}, ScreenSpot~\cite{cheng2024screenspot}, ScreenSpot-v2~\cite{wu2024screenspotv2} and ScreenSpot-Pro~\cite{li2025screenspotpro}.
    \item \href{https://github.com/Espere-1119-Song}{Espere-1119-Song} (Enxin Song): The contributor helped support Video-MMLU~\cite{song2025videomlu}, MovieChat1k~\cite{song2024moviechat} and VDC~\cite{chai2024auroracap}.
    \item \href{https://github.com/maosong2022}{maosong2022} (Song Mao): The contributor helped support MMVP~\cite{zhang2024mmvp}, CVBench~\cite{tong2024cambrian} and CharXiv~\cite{wang2024charxiv}, who also help solved many issues in the community.
    \item \href{https://github.com/SYuan03}{SYuan03} (Shengyuan Ding): The contributor supports mPLUG-Owl3~\cite{ye2024mplug}, MM-IFEval~\cite{ding2025mmif}, ChartMimic~\cite{yang2024chartmimic} and fixed problems in Creation-MMBench~\cite{fang2025creation} evaluation.
    \item \href{https://github.com/TianhaoLiang2000}{TianhaoLiang2000} (Tianhao Liang): The contributor supports mega-bench~\cite{chen2024mega}, Kimi-VL-A3B~\cite{team2025kimi} and add vllm support for QwenVL/LLama4/InternVL Series Model.
    \item \href{https://github.com/zzc-1998}{zzc-1998} (Zicheng Zhang): The contributor supports Q-Bench~\cite{wu2023q}, A-Bench~\cite{zhang2024bench} and Q-Bench Video~\cite{zhang2025q}.
\end{itemize}

\textbf{Qualified Contributors (2026.04):}
\begin{itemize}[itemsep=0pt,topsep=0pt,parsep=0pt]
    \item \href{https://github.com/Bujiazi}{Bujiazi} (Jiazi Bu): The contributor helped support ChartQAPro, ChartMuseum, and SimpleVQA.
    \item \href{https://github.com/mjuicem}{mjuicem} (Junming Lin): The contributor helped support RefCOCO and DREAM-1K, and added support for the Qwen3-VL series.
    \item \href{https://github.com/Liuziyu77}{Liuziyu77} (Ziyu Liu): The contributor helped support olmOCRBench, OceanOCRBench, and MAT-Bench in a single OCR benchmark bundle.
    \item \href{https://github.com/JoeLeelyf}{JoeLeelyf} (Yifei Li): The contributor helped support UniSVG, SArena, and SArena-MINI, including a follow-up bug fix for SArena-MINI.
    \item \href{https://github.com/Li-Jinsong}{Li-Jinsong} (Jinsong Li): The contributor helped support PlotQA, ChartX, and ChartBench, and improved the robustness of ChartX evaluation.
    \item \href{https://github.com/Wiselnn570}{Wiselnn570} (Xilin Wei): The contributor helped support WorldVQA, VisualPuzzles, and PuzzleVQA in a single bundled PR.
    \item \href{https://github.com/SunzeY}{SunzeY} (Zeyi Sun): The contributor helped support OSWorld-G, VBGD, and VenusBench-GD, building out the GUI-grounding benchmark suite.
    \item \href{https://github.com/Qi-Zhangyang}{Qi-Zhangyang} (Zhangyang Qi): The contributor helped support DA-2K, ERQA, and RefSpatialBench, contributing the bulk of the spatial-grounding additions.
    \item \href{https://github.com/mzr1996}{mzr1996} (Zerun Ma): The contributor helped support MaCBench and VideoMMMU, fixed evaluation timeouts for ChartMimic, parallelized UniSVG evaluation, fixed Physics and MM-IFEval, and updated model settings for the 2026.02 live leaderboard.
    \item \href{https://github.com/zhaowei-wang-nlp}{zhaowei-wang-nlp} (Zhaowei Wang): The contributor helped support mmlongbench, memlens, and longdocurl.
    \item \href{https://github.com/Junjun2016}{Junjun2016} (Junjun He): The contributor helped support and maintain medical-related benchmarks.
\end{itemize}

\subsection{Full Contributor List}

Report co-authors are excluded from the below list. 

\begin{itemize}[itemsep=0pt,topsep=0pt,parsep=0pt]
    \item \href{https://github.com/echo840}{echo840}: The contributor supports OCRBench~\cite{liu2023hidden}.
    \item \href{https://github.com/tianyu-z}{tianyu-z}, \href{https://github.com/sheryc}{sheryc}: The contributor supports VCR~\cite{zhang2024vcr}.
    \item \href{https://github.com/TousenKaname}{TousenKaname}: The contributor supports GMAI-MMBench~\cite{chen2024gmai} and MGM-7B~\cite{li2024mini}.
    \item \href{https://github.com/ShuoZhang2003}{ShuoZhang2003}: The contributor supports Monkey~\cite{li2023monkey}.
    \item \href{https://github.com/PCIResearch}{PCIResearch}: The contributor supports TransCoreM.
    \item \href{https://github.com/dylanqyuan}{dylanqyuan}: The contributor supports AesBench~\cite{huang2024aesbench}.
    \item \href{https://github.com/mary-0830}{mary-0830}: The contributor supports OmChat~\cite{zhao2024omchat} and VLM-R1~\cite{shen2025vlmr1}.
    \item \href{https://github.com/LZHgrla}{LZHgrla}: The contributor supports LLaVA-XTuner~\cite{2023xtuner}.
    \item \href{https://github.com/bingwork}{bingwork}: The contributor supports MMAlaya and MMAlaya2~\cite{datacanvas2024mmalaya}.
    \item \href{https://github.com/fitzpchao}{fitzpchao}: The contributor supports ShareCaptioner~\cite{chen2023sharegpt4v} and CogVLM~\cite{Wang2023CogVLMVE}.
    \item \href{https://github.com/Isaachhh}{Isaachhh}: The contributor adds custom prompts for Bunny~\cite{he2024bunny}.
    \item \href{https://github.com/eltociear}{eltociear}: The contributor adds the Japanese README.
    \item \href{https://github.com/iyuge2}{iyuge2}: The contributor supports GLM-Vision and updates prompts for CogVLM \& GLM4v-9B.
    \item \href{https://github.com/BrenchCC}{BrenchCC}: The contributor supports Mantis~\cite{Jiang2024MANTISIM}.
    \item \href{https://github.com/Ezra-Yu}{Ezra-Yu}: The contributor fixes an error in the acc calculation.
    \item \href{https://github.com/StarCycle}{StarCycle}: The contributor supports local LLMs as the judge.
    \item \href{https://github.com/shenyunhang}{shenyunhang}: The contributor fixes a video inference bug.
    \item \href{https://github.com/HugoLaurencon}{HugoLaurencon}: The contributor fixes Idefics2 MathVista prompt.
    \item \href{https://github.com/lerogo}{lerogo}: The contributor converts base64 images to memory.
    \item \href{https://github.com/lihytotoro}{lihytotoro}: The contributor supports MiniCPM-V-2.6 and MiniCPM-o-2.6~\cite{yao2024minicpm}.
    \item \href{https://github.com/hkunzhe}{hkunzhe}: The contributor supports TableVQA~\cite{kim2024tablevqa}.
    \item \href{https://github.com/azshue}{azshue}: The contributor fixes an XGen-MM problem.
    \item \href{https://github.com/anzhao920}{anzhao920}: The contributor supports RBDash.
    \item \href{https://github.com/youngfly11}{youngfly11}: The contributor supports CORE-MM~\cite{han2023infimm}.
    \item \href{https://github.com/YuZhiyin}{YuZhiyin}: The contributor supports Claude3-V.
    \item \href{https://github.com/VictorSanh}{VictorSanh}: The contributor adds Idefics2 custom prompts.
    \item \href{https://github.com/Jize-W}{Jize-W}: The contributor fixes an XComposer inference error.
    \item \href{https://github.com/YJY123}{YJY123}: The contributor supports XVERSE-V-13B.
    \item \href{https://github.com/naoto0804}{naoto0804}: The contributor supports Azure OpenAI API.
    \item \href{https://github.com/binwang777}{binwang777}: The contributor supports 360VL-70B.
    \item \href{https://github.com/KainingYing}{KainingYing}: The contributor supports MMT-Bench~\cite{ying2024mmt}.
    \item \href{https://github.com/FeipengMa6}{FeipengMa6}: The contributor supports WeMM~\cite{wemm} and WeThink-Qwen2.5VL-7B~\cite{yang2025wethink}.
    \item \href{https://github.com/scikkk}{scikkk}: The contributor supports Math-Vision~\cite{wang2025measuring}.
    \item \href{https://github.com/zzc-1998}{zzc-1998}: The contributor supports Q-Bench~\cite{wu2023q} and A-Bench~\cite{zhang2024bench}.
    \item \href{https://github.com/weikaih04}{weikaih04}: The contributor supports TaskMeAnything-V1-ImageQA-Random~\cite{zhang2024task}.
    \item \href{https://github.com/kq-chen}{kq-chen}: The contributor supports Qwen2-VL~\cite{wang2024qwen2}.
    \item \href{https://github.com/zeyofu}{zeyofu}: The contributor supports BLINK~\cite{fu2024blink}.
    \item \href{https://github.com/jinyu121}{jinyu121}: The contributor adds the TextMCQ dataset class.
    \item \href{https://github.com/Quakumei}{Quakumei}: The contributor fixes a README typo.
    \item \href{https://github.com/max-yue}{max-yue}: The contributor fixes a README typo.
    \item \href{https://github.com/dongyh20}{dongyh20}: The contributor of Ola~\cite{liu2025ola}.
    \item \href{https://github.com/lrlbbzl}{lrlbbzl}: The contributor of URSA-8B and URSA-8B-PS-GRPO~\cite{luo2025ursa}.
    \item \href{https://github.com/lyccnb}{lyccnb}: The contributors of CG-Bench~\cite{chen2024cg}.
    \item \href{https://github.com/jscslld}{jscslld}: The contributors of CG-Bench~\cite{chen2024cg} and CG-AV-Counting~\cite{lu2025av}.
    \item \href{https://github.com/andrewliao11}{andrewliao11}: The contributor of Q-Spatial Bench~\cite{liao2024reasoning}.
    \item \href{https://github.com/wulipc}{wulipc}: The contributor of CC-OCR~\cite{yang2024cc}.
    \item \href{https://github.com/mfarre}{mfarre}: The contributor of SmolVLM and SmolVLM2.
    \item \href{https://github.com/ttguoguo3}{ttguoguo3}: The contributor of CRPE~\cite{wang2023allseeing,wang2024allseeing_v2}, MMNIAHBench~\cite{wang2025needle}.
    \item \href{https://github.com/teowu}{teowu}, \href{https://github.com/Coobiw}{Coobiw}: The contributors of Aria~\cite{aria}.
    \item \href{https://github.com/white2018}{white2018}: The contributor of MiniMonkey~\cite{huang2024mini}.
    \item \href{https://github.com/LingyiHongfd}{LingyiHongfd}: The contributor of WorldSense~\cite{hong2025worldsense}.
    \item \href{https://github.com/Khang-9966}{Khang-9966}, \href{https://github.com/huynhbaobk}{huynhbaobk}: The contributors of VIntern~\cite{doan2024vintern1befficientmultimodallarge}.
    \item \href{https://github.com/ZhangYuanhan-AI}{ZhangYuanhan-AI}: The contributor of LLaVA-Video~\cite{zhang2024video}.
    \item \href{https://github.com/CMeteor}{CMeteor}: The contributor of TeleMM.
    \item \href{https://github.com/CaraJ7}{CaraJ7}: The contributor of MathVerse~\cite{zhang2024mathverse} and MME-COT~\cite{jiang2025mmecot}.
    \item \href{https://github.com/jiutiancv}{jiutiancv}: The contributor of JT-VL-Chat.
    \item \href{https://github.com/Baiqi-Li}{Baiqi-Li}: The contributor of NaturalBench~\cite{li2025naturalbench}.
    \item \href{https://github.com/DataWizardLiu}{DataWizardLiu}: The contributor of DoubaoVL.
    \item \href{https://github.com/TobiasLee}{TobiasLee}: The contributor of VLRewardBench~\cite{li2024vlrewardbench}.
    \item \href{https://github.com/ZhiminYao1}{ZhiminYao1}: The contributor of Taiyi.
    \item \href{https://github.com/Haochen-Wang409}{Haochen-Wang409}: The contributor of Ross~\cite{wang2024reconstructive}.
    \item \href{https://github.com/hills-code}{hills-code}: The contributor of Janus-1.3B~\cite{wu2024janus}.
    \item \href{https://github.com/nbl97}{nbl97}: The contributor of HunYuan API.
    \item \href{https://github.com/ChuanyangZheng}{ChuanyangZheng}: The contributor of BailingMM API.
    \item \href{https://github.com/lcysyzxdxc}{lcysyzxdxc}: The contributor of R-Bench~\cite{li2024r}.
    \item \href{https://github.com/lerogolerogo}{lerogolerogo}: The contributor of MMGenBench~\cite{huang2024mmgenbench}.
    \item \href{https://github.com/thomas-yanxin}{thomas-yanxin}: The contributor of XinYuan-VL-2B.
    \item \href{https://github.com/andimarafioti}{andimarafioti}: The contributor of SmolVLM-256B/500B.
    \item \href{https://github.com/wufeim}{wufeim}: The contributor of 3DSRBench~\cite{ma20243dsrbench}.
    \item \href{https://github.com/yangyue5114}{yangyue5114}: The contributor of MMVet-Hard.
    \item \href{https://github.com/gushu333}{gushu333}: The contributor of Aquila-VL-2B~\cite{gu2024infinitymmscalingmultimodalperformance}.
    \item \href{https://github.com/bobo0810}{bobo0810}: The contributor of Phi-4~\cite{abdin2024phi4}.
    \item \href{https://github.com/cmatachuan}{cmatachuan}: The contributor of SAIL-VL-1.5 and SAIL-VL-1.6~\cite{dong2025sail}.
    \item \href{https://github.com/dellixx}{dellixx}: The contributor of Janus-Pro-1B~\cite{chen2025januspro}.
    \item \href{https://github.com/g-h-chen}{g-h-chen}: The contributor of UCSC-VLAA-Thinker~\cite{chen2025vlaathinker}.
    \item \href{https://github.com/maojialiang}{maojialiang}: The contributor of Ristretto.
    \item \href{https://github.com/RainJamesY}{RainJamesY}: The contributor of VLM2Bench~\cite{zhang2025vlm2bench}.
    \item \href{https://github.com/ryf1123}{ryf1123}: The contributor of VGRP-Bench~\cite{ren2025vgrp}.
    \item \href{https://github.com/suencgo}{suencgo}: The contributor of Physics~\cite{feng2025physics}.
    \item \href{https://github.com/suyccc}{suyccc}: The contributor of VMCBench~\cite{zhang2025vmcbench}.
    \item \href{https://github.com/sync-yxh}{sync-yxh}: The contributor of Taichu-VLR.
    \item \href{https://github.com/tangkexian}{tangkexian}: The contributor of LEGO-Puzzle~\cite{tang2025lego}.
    \item \href{https://github.com/tianbinli}{tianbinli}: The contributor of MedXpertQA~\cite{zuo2025medxpertqa}.
    \item \href{https://github.com/waltsun}{waltsun}: The contributor of MOAT~\cite{ye2025moat}.
    \item \href{https://github.com/weiyao-wang}{weiyao-wang}: The contributor of AKI~\cite{wang2025AKI}.
    \item \href{https://github.com/xingruiwang}{xingruiwang}: The contributor of spatial457~\cite{wang2025spatial457}.
    \item \href{https://github.com/xwy-bit}{xwy-bit}: The contributor of VisuLogic~\cite{xu2025visulogic}.
    \item \href{https://github.com/yangtian6781}{yangtian6781}: The contributor of MMCR~\cite{tian2025mmcr}.
    \item \href{https://github.com/zhaomh1998}{zhaomh1998}: The contributor of TDBench~\cite{hou2025tdbench}.
    \item \href{https://github.com/Hyggge}{Hyggge}: The contributor of valley2~\cite{wu2025valley2}.
    \item \href{https://github.com/xjtupanda}{xjtupanda}: The contributor of $V^*$ Benchmark~\cite{wu2024vstar}.
    \item \href{https://github.com/wutaiqiang}{wutaiqiang}: The contributor of PhyX~\cite{shen2025phyx}.
    \item \href{https://github.com/mxin262}{mxin262}: The contributor of OCR-Reasoning~\cite{huang2025ocrreasoning}.
    \item \href{https://github.com/Zhouzone}{Zhouzone}: The contributor of MSEarth~\cite{zhao2025msearth} and OmniEarth~\cite{wang2025omniearth}.
    \item \href{https://github.com/An-LanWang}{An-LanWang}: The contributor of WildDoc~\cite{wang2025wilddoc}.
    \item \href{https://github.com/JunyingWang959}{JunyingWang959}: The contributor of A4Bench~\cite{wang2025affordance}.
    \item \href{https://github.com/JiakangYuan}{JiakangYuan}: The contributor of MME-Reasoning~\cite{yuan2025mmerea}.
    \item \href{https://github.com/donahowe}{donahowe}: The contributor of Video-Holmes~\cite{cheng2025videoholmes}.
    \item \href{https://github.com/rohunagrawal}{rohunagrawal}: The contributor of Omni3D-Bench~\cite{marsili2025omni3d}.
    \item \href{https://github.com/snowclipsed}{snowclipsed}: The contributor of TallyQA Dataset~\cite{acharya2019tallyqa}.
    \item \href{https://github.com/JonasLoos}{JonasLoos}: The contributor of SCAM Dataset~\cite{westerhoff2025scam}.
    \item \href{https://github.com/zjh-tsinghua}{zjh-tsinghua}: The contributor of FlashVL~\cite{zhang2025flash}.
    \item \href{https://github.com/nanocm}{nanocm}: The contributor of OmniEarth~\cite{wang2025omniearth} and XLRS-Bench-Lite~\cite{wang2025xlrs}.
    \item \href{https://github.com/zhouyiks}{zhouyiks}: The contributor of MMVMBench
    \item \href{https://github.com/aarbelle}{aarbelle}: The contributor of GraniteVision~\cite{team2025granite}.
    \item \href{https://github.com/yuzeng0-0}{yuzeng0-0}: The contributor of VCR Bench~\cite{qi2025vcr}.
    \item \href{https://github.com/Amber0614}{Amber0614}: The contributor of GOBench.
    \item \href{https://github.com/timothycdc}{timothycdc}: The contributor of Aya Vision Bench.
    \item \href{https://github.com/ignoreandfly}{ignoreandfly}: The contributor of TopViewRS.
    \item \href{https://github.com/liujiyaoFDU}{liujiyaoFDU}: The contributor of MedQ-Bench.
    \item \href{https://github.com/HengjunPu}{HengjunPu}: The contributor fixes API-based evaluation for OlympiadBench.
    \item \href{https://github.com/PengDa02}{PengDa02}: The contributor fixes MLVU MCQ ordering issues.
    \item \href{https://github.com/Pig255}{Pig255}: The contributor fixes the relaxed-correctness parameter ordering.
    \item \href{https://github.com/ffanyt}{ffanyt}: The contributor fixes evaluation issues for BMMR.
    \item \href{https://github.com/PeterWangyi}{PeterWangyi}: The contributor adds a generic prepare\_tsv method to the VideoBaseDataset, fixes single-image PNG saving and 3DSR evaluation, and helps merge EASI-added benchmarks and models.
    \item \href{https://github.com/MarkovChain-why}{MarkovChain-why}: The contributor adds support for additional dataset evaluation cases.
    \item \href{https://github.com/isjinghao}{isjinghao}: The contributor extends multimodal-medical evaluation cases.
    \item \href{https://github.com/anmi9508}{anmi9508}: The contributor of Logics-Thinking-32B and Logics\_Thinking models.
    \item \href{https://github.com/floe20}{floe20}: The contributor of MiniCPM-V-4 and MiniCPM-V-4.5.
    \item \href{https://github.com/Ella-chanw}{Ella-chanw}: The contributor of GLM-4.5V and GLM-4.1V inference updates.
    \item \href{https://github.com/hi-wesley}{hi-wesley}: The contributor fixes the Gemini API backend routing.
    
\end{itemize}